# An agentic generative large language model for treatment planning of colorectal cancer

**Authors:** Mengxian Lyu[1]†, Cheng Peng[1]†, Tim Jang[7,12], Ang Li[1], Mengyuan Zhang[1], Ziyi Chen[1], Leighton Elliott[7,12], Tianshi Liu[7,12], Lidice Galindo[7,12], Chiranjeevi Sainatham[7,12], Oscar F. Borja-Montes[7,12], Kaleb E. Smith[3], Ying Zhang[4], Lichao Sun[9], Jiang Bian[10, 11], Gloria Lipori[5, 6], Duane A. Mitchell[6,12], Elizabeth A Shenkman[1,12], Yi Guo[1,2,12], Thomas J. George[7,12], Yonghui Wu[1,2,12 *]

**Affiliations:**

[1]Department of Health Outcomes and Biomedical Informatics, College of Medicine, University of Florida, Gainesville, Florida, USA.

[2]Preston A. Wells, Jr. Center for Brain Tumor Therapy, Lillian S. Wells Department of Neurosurgery, University of Florida, Gainesville, Florida, USA

[3]NVIDIA, Santa Clara, California, USA.

[4]Research Computing, University of Florida, Gainesville, Florida, USA.

[5]Integrated Data Repository Research Services, University of Florida, Gainesville, Florida, USA.

[6]Lillian S. Wells Department of Neurosurgery, UF Clinical and Translational Science Institute, University of Florida, Gainesville, FL, USA

[7]Division of Hematology & Oncology, Department of Medicine, College of Medicine, University of Florida, Gainesville, FL, USA

[8]Division of Endocrinology, Department of Medicine, College of Medicine, University of Florida, Gainesville, FL, USA

[9]Department of Computer Science and Engineering, Lehigh University, Bethlehem, PA, USA

[10]Department of Biostatistics and Health Data Science, School of Medicine, Indiana University, Indianapolis, IN, USA

[11]Regenstrief Institute, Indianapolis, IN, USA

[12]University of Florida Health Cancer Institute, Gainesville, FL, USA

† These authors contributed equally to this work.

*Corresponding author

Yonghui Wu, PhD

The Malachowsky Data Science & Information Technology Building

1889 Museum Rd, 7th Floor, Suite 7000

Gainesville, FL, USA, 32611

Phone: 352-294-8436

Email: yonghui.wu@ufl.edu

**Abstract**

Treatment planning in precision oncology requires synthesizing heterogeneous patient information with rapidly evolving clinical guidelines to ensure guideline-concordant care. While large language models (LLMs) show promise in many diagnostic tasks, their adoption for high-stakes treatment planning is hindered by complex reasoning, adherence to timely clinical guidelines, and safety concerns. In this study, we present GatorOnco, an agentic LLM for colorectal cancer (CRC) treatment planning. GatorOnco is developed using a total of 282 billion tokens of biomedical text, including a healthcare system-scale clinical text with 166 billion tokens from the University of Florida (UF) Health. We implemented a domain-adaptation method that integrates pre-training, model merging, a two-stage post-training approach, and agent-based reinforcement learning. An agentic retrieval-augmented generation (RAG) approach is used to dynamically integrate time-sensitive clinical guidelines into the reasoning process. In a blind, randomized clinical evaluation conducted by five UF Health oncologists, GatorOnco significantly outperformed ($P < 0.01$) open-source LLMs and achieved expert-level performance on par with UF Health oncologists. Specifically, compared with expert oncologists, GatorOnco was rated significantly higher for readability (4.46 versus 4.19, $P < 0.01$) and completeness (3.91 versus 3.52, p = <0.01); was rated statistically comparable (i.e., on par) for correctness (4.09 versus 4.11, p = 0.921), currency (4.04 versus 3.98, p = 0.478), and safety (4.22 versus 4.22, p = 0.999). Our findings demonstrate that integrating agentic reasoning with large-scale domain adaptation can bridge the gap of generative AI in high-stakes cancer treatment planning.

**Introduction**

Treatment planning is complex yet crucial for better treatment outcomes in cancer care[1,2]. Oncologists must synthesize complex information from disease biology, prior therapies and toxicities, comorbidities, biomarker profiles, and contemporary clinical guidelines to provide personalized therapy that optimizes treatment benefit. Despite progress in clinical decision support systems[3], much of the treatment planning remains manual and time-intensive, challenging consistent high-quality treatment of cancer and jeopardizing timely, guideline-concordant cancer care[4]. In colorectal cancer, the third leading cause of cancer death, approximately 25% of patients received guideline-discordant initial therapy[5], which was associated with higher healthcare costs and utilization[6]. This discordance is exacerbated by the rapid development of new cancer treatments and the high frequency of refinements to the standard of care. The National Comprehensive Cancer Network® (NCCN®) guidelines, for instance, underwent 241 updates across 88 guidelines in 2024[7].

Conventional clinical decision support systems for treatment planning are predominantly based on deterministic templates and heuristic rules, which fail to capture personalized nuances and lack expert-level clinical reasoning in precision oncology[8]. While LLMs have demonstrated utility in many tasks—such as patient information extraction[9], clinical documentation[10,12], disease screening[11], and differential diagnosis[12,13]—their application in treatment planning is limited. General-purpose LLMs have been reported to have a high propensity for guideline-discordant and clinical safety[14], potentially caused by gaps in clinical reasoning, grounding to clinical factuality, and integration with timely clinical guidelines.[15] While advanced learning algorithms such as reinforcement learning (RL) have endowed LLMs with robust reasoning capabilities in mathematics and computer coding[16], translating this "reasoning" to oncology presents critical challenges[17]. Specifically, current models are reported to struggle with catastrophic forgetting during domain adaptation and lack the autonomous planning capacity to integrate time-sensitive clinical guidelines, posing risks to safe, guideline-concordant, expert-level therapeutic recommendations.

To bridge this gap, we propose a domain-adaptation method and develop GatorOnco, an agentic-based generative LLM specialized for CRC treatment planning. GatorOnco is developed using a healthcare-system-scale corpus of 282 billion tokens—including 166 billion tokens of longitudinal clinical text from the University of Florida (UF) Health. UF Health oncologists developed a treatment-planning dataset using data from 395 real-world UF Health CRC patients. We propose a novel domain-adaptation method that integrates pre-training, model merging, a two-stage post-training, and agent-based reinforcement learning. We designed a hierarchical reward function that strictly prioritizes therapeutic correctness and patient safety and an agentic retrieval-augmented generation (RAG) to function as an autonomous clinical reasoning engine. As an agentic AI, GatorOnco can proactively navigate up-to-date NCCN guidelines, dynamically formulate subqueries, digest evidence for CRC patients, and synthesize guideline-concordant treatment plans. In a blind, randomized test involving 79 real-world UF Health CRC patients and UF Health oncologists, GatorOnco significantly outperformed open-source baselines ($P < 0.01$) and achieved expert-level performance, demonstrating safety and correctness profiles statistically indistinguishable from UF Health oncologists.

## Results

### Development and architecture of GatorOnco

We developed GatorOnco through a multi-stage training strategy (Fig. 1a). Starting from a widely used open-source LLM, Meta-Llama-3.1-8B[18], we performed continued pre-training using a 282-billion-token biomedical corpus, including 166 billion tokens of clinical text from UF Health, to create GatorTronLlama (training and validation loss curves are shown in Supplementary Fig. 1). To mitigate catastrophic forgetting and preserve instruction-following capabilities, we merged GatorTronLlama with its instruction-tuned Llama model using model merging (mergekit-evo[19]). We systematically compared multiple merge strategies and selected the best-performing method based on balanced performance in clinical knowledge benchmarks and instruction-following capabilities.

UF Health oncologists identified and manually annotated a cohort of 395 CRC cases, which were split into 316 cases for model development (276 for training, 40 for validation), and 79 held-out test cases. We applied supervised fine-tuning (SFT) followed by reinforcement learning (RL) to align the merged model with expert clinical reasoning. Specifically, during RL, we developed an agentic training algorithm with interleaved retrieval to ground treatment plan generation in contemporary NCCN guidelines (Colon： v4.2025; Rectal: v3.2025) and enforce guideline currency during optimization. The instructed model, GatorOnco, is an agentic generative AI specialized for CRC treatment planning. **Fig. 1** shows an overview of the study design and workflow.

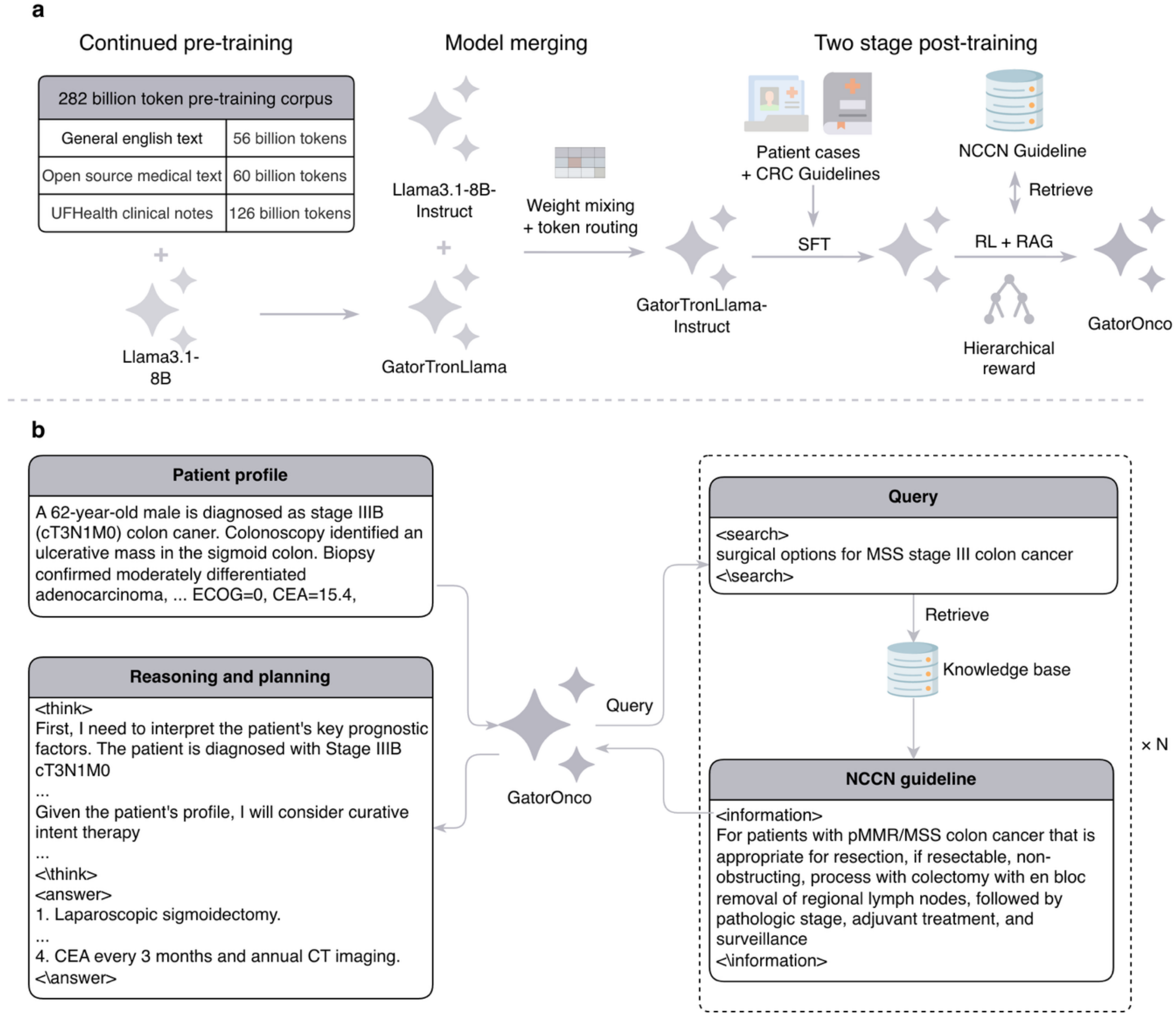

**Fig. 1: Development and workflow of GatorOnco. a,** Continued pre-training, model merging, and RL using CRC cases and NCCN guidelines. SFT, Supervised fine-tuning; RL, Reinforcement learning; RAG, Retrieval Augment Generation; NCCN, National Comprehensive Cancer Network. **b,** Agentic-based interleaved retrieval-generation workflow that incorporates NCCN evidence into explicit reasoning to produce guideline-concordant treatment plans.

## UF Health CRC patient cases

We identified a total of 3,915 CRC patients in the UF Health Integrated Data Repository (IDR), which captures longitudinal EHRs since 2012. After deduplication, we applied a multi-stage filtering process (Fig. 2a) to identify clinical notes from the initial oncologist encounters, where treatment plans are primarily documented. Three UF Health oncologists subsequently reviewed and selected patient cases for further annotation. Oncologists categorized the therapies into 5 categories (surgery, radiotherapy, chemotherapy, targeted therapy, and palliative care) and manually reviewed each cases to annotate (1) the therapy provided to the patient, (2) all potential therapies that can be considered for the patients per experts judgements, and (3) the narrative descriptions of treatment plan and the evidence used for treatment planning The final dataset has 395 CRC cases, which were divided into a training set of 316 cases for developing AI models and a testing set of 79 cases for evaluation using stratified sampling based on age, sex, primary tumor site, and AJCC 7th stage. As shown in Table 1 and Fig. 2b, there are no statistically significant differences between the training and test sets.

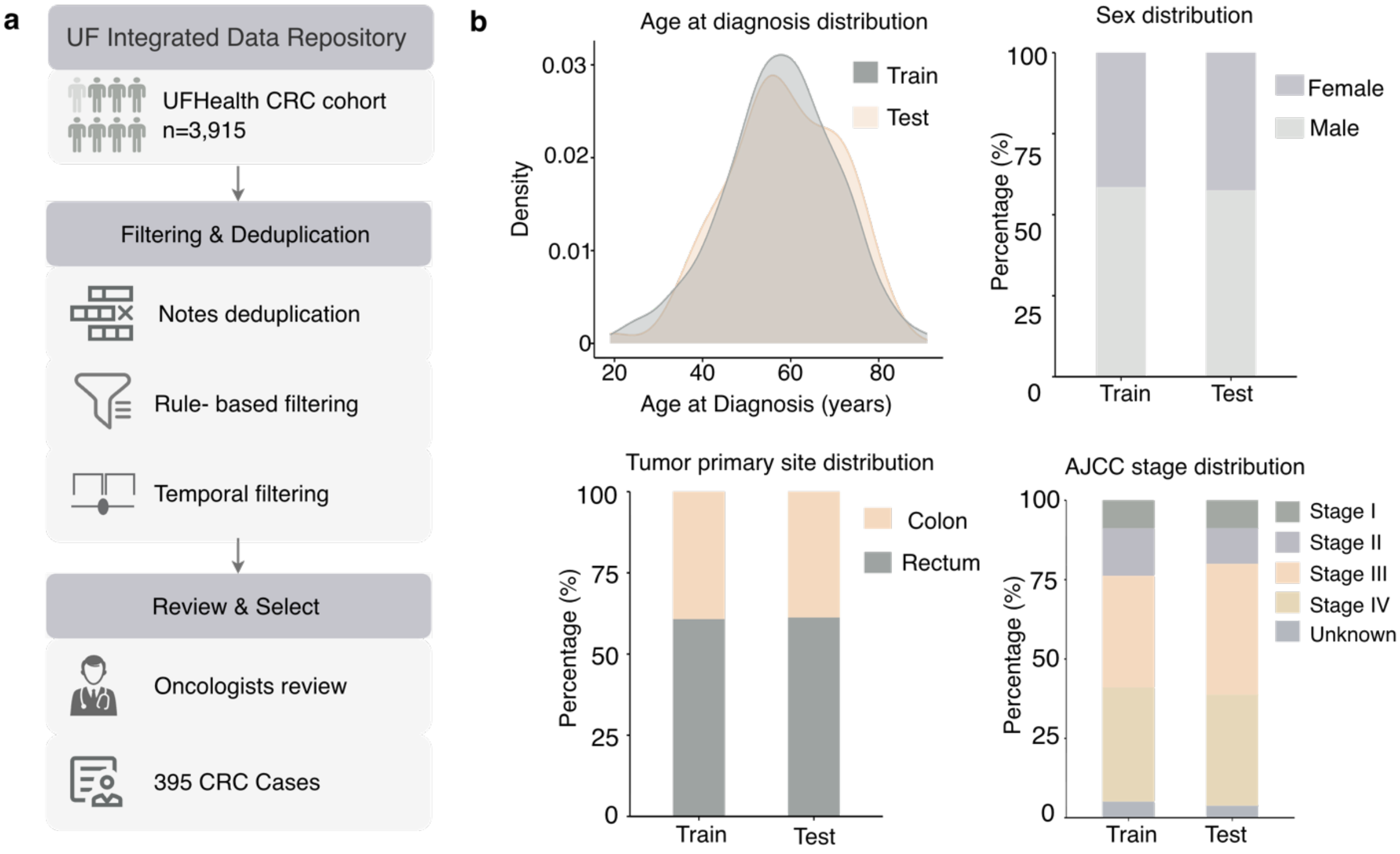

**Fig. 2: Oncologists review CRC cases to create ground-truth for training and evaluation. a,** Multi-stage filtering and expert review to construct a ground-truth dataset of 395 CRC cases from the UF Health**. b,** Comparison of baseline characteristics (age, sex, tumor site, AJCC stage) between the training set and the test set.

**Table 1. Demographic and clinical characteristics of training and test sets.**

| Characteristic | Train (N = 316) | Test (N = 79) | p-value |
|---|---|---|---|
| **Age at diagnosis, years** | 57.7 ± 12.9 | 58.3 ± 12.7 | 0.735 |
| **Sex, n (%)** | | | 0.959 |
| Female | 131 (41.5%) | 33 (41.8%) | |
| Male | 185 (58.5%) | 46 (58.2%) | |
| **Race, n (%)** | | | 0.638 |
| Asian | 1 (0.3%) | 1 (1.3%) | |
| Black or African American | 53 (16.8%) | 11 (13.9%) | |
| White | 250 (79.1%) | 63 (79.7%) | |
| Other / Unknown | 12 (3.8%) | 4(5.1%) | |
| **AJCC 7th stage, n (%)** | | | 0.746 |
| Stage I | 28 (8.9%) | 7 (8.9%) | |
| Stage II | 50 (15.8%) | 9 (11.4%) | |
| Stage III | 107 (33.8%) | 32 (40.5%) | |
| Stage IV | 114 (36.1%) | 28 (35.4%) | |
| Unknown | 17 (5.4%) | 3 (3.8%) | |
| **Primary tumor site, n (%)** | | | 1.000 |
| Colon | 124 (39.2%) | 31 (39.2%) | |
| Rectum | 192 (60.8%) | 48 (60.8%) | |

Continuous variables are presented as mean ± standard deviation and compared using two-sample t-tests. Categorical variables are presented as counts and percentages and compared using $\chi^2$ tests. All p-values reflect two-sided statistical tests.

**Model merging, supervised fine-tuning, and reinforcement learning with clinically oriented rewards**

We applied model merging to merge GatorTronLlama with Llama-3.1-8B-Instruct via weight mixing and token routing[20], generating a new model, GatorTronLlama-Instruct, which inherited the instruction-following ability from Llama-3.1-8B-Instruct and inherited the clinical text generation ability from GatorTronLlama. Then, the merged model was fine-tuned using a total of 6.8k question–answer pairs distilled from the NCCN guidelines, 23.5k case report-style clinical reasoning corpus, and the 316 CRC cases in the training set (see Supplement for details). We applied agentic reinforcement learning (RL) with interleaved retrieval over the NCCN guidelines, enabling the model to initiate search actions during generation and use retrieved guidelines to guide treatment plan generation through reasoning. RL was optimized using Group Relative Policy Optimizatio[21] (GRPO) under a KL-divergence constraint relative to the reference model. To ensure AI safety, we collaborated with UF Health oncologists and co-designed a "Safety-Oriented" Hierarchical Reward Function. This reward function enforces that our model only generates specific treatment regimens (e.g., FOLFOX) *only if* it correctly identifies the high-level therapeutic modality (e.g., Chemotherapy). By integrating this clinically grounded signal into the Group Relative Policy Optimization (GRPO) framework, we seek to enforce a safety-oriented policy and alleviate potential hallucinations using the widely adopted Group Relative Policy Optimization (GRPO) framework. (see Methods for mathematical formulation).

**GatorOnco determines categorical CRC treatment options**

We evaluate GatorOnco for single-therapy determination - to determine the therapy delivered to CRC patients after encounter with oncologists, and multi-therapy determination – to determine potential therapies that can be considered for the patient before the encounter, a common treatment planning task for oncologists. We compare GatorOnco with 2 widely used general-purpose LLMs, including Llama-3.1-8B-Instruct, Llama-3.3-70B-Instruct[18], and a widely used medical LLM, MedGemma-27B-Text-it[22]. As shown in Figure 3.a, for single-therapy determination, GatorOnco achieved the best macro-F1 score of 0.924, remarkably higher than other models, which had F1 scores from 0.748 to 0.910. Through pre-training on healthcare system-scale EHRs and post-training, including supervised fine-tuning and reinforcement learning, GatorOnco outperformed Llama-3.3-70B, a general-domain LLM that is almost 9 times larger. For high-stakes invasive treatments, including surgery and chemotherapy, GatorOnco demonstrated much higher performance than the conservative treatments. (Fig. 3b) GatorOnco achieved F1-scores of 0.979 for chemotherapy, 0.964 for surgery, and 0.933 for radiotherapy, significantly outperforming the much larger Llama-3.3-70B.

We further evaluated GatorOnco to determine multiple therapies that can be considered for CRC patients, a common routine task for oncologists, particularly in the context of shared decision-making with patients and their caregivers (Fig. 3c). GatorOnco achieved the best F1 score of 0.913, outperforming other generative LLMs. GatorOnco suggested the correct multiple-therapy options for 73 of 79 cases (92.4%), compared with 55–68 cases (69.6–86.1%) by other LLMs (Fig. 3d).

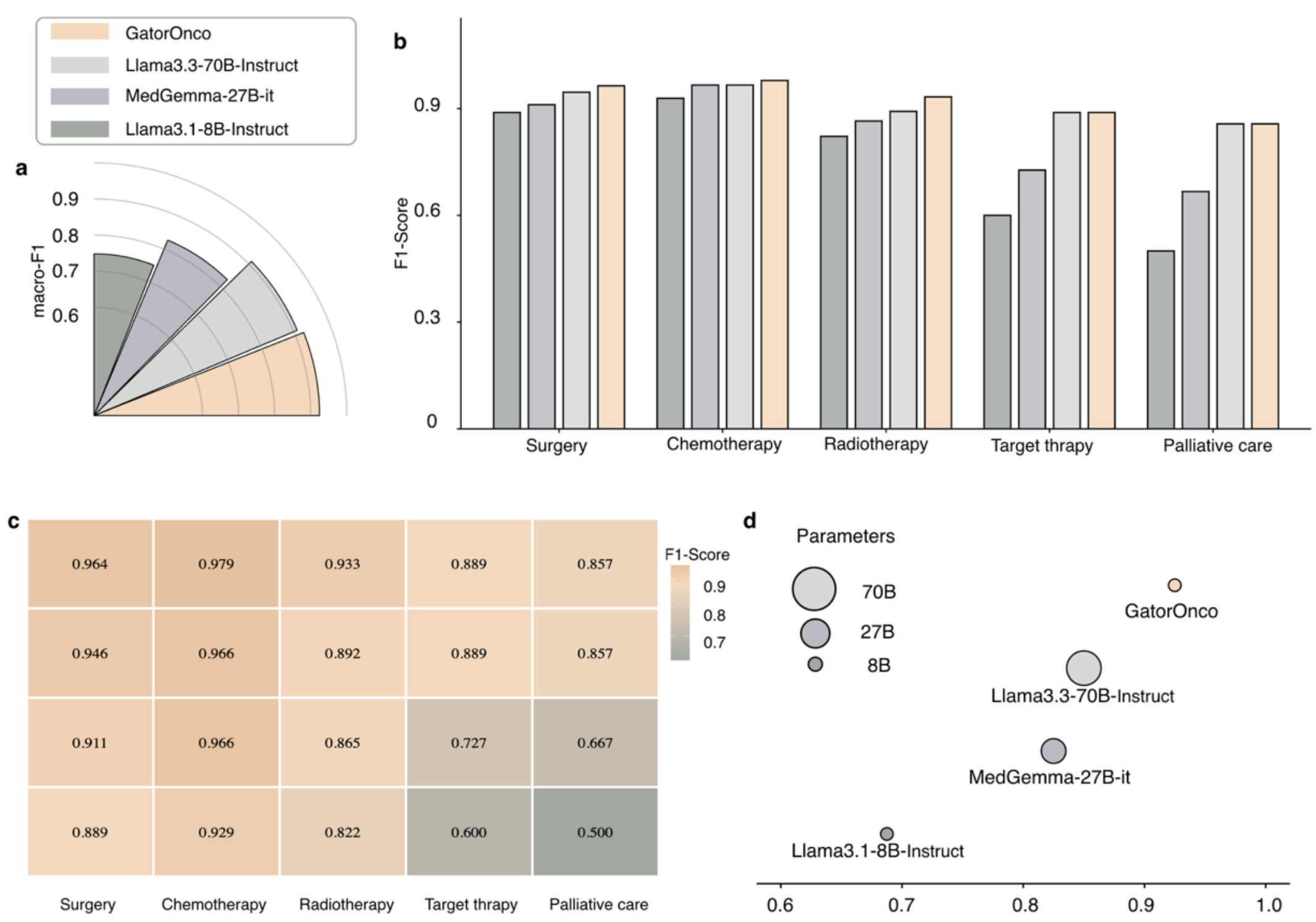

**Fig. 3: Evaluation of treatment option determination. a,** Macro-average F1-Score across six treatment options. **b,** Per-option F1-Score for single-label treatment determination. **c,** treatment option set determination F1-Score. Rows from top to bottom represent: GatorOnco, Llama3.3-70B-Instruct, MedGemma-27B-it, and Llama3.1-8B-Instruct. **d,** Exact-match accuracy for treatment determination, with bubble size proportional to model parameter size.

## GatorOnco generates narrative treatment plans

We applied GatorOnco to digest heterogeneous patient information and generate narrative treatment plans. We evaluated GatorOnco-generated treatment plans against expert-written plans as ground truth. Following prior work, we used automatic evaluation metrics to assess lexical fidelity, semantic correspondence, and factual grounding, and we report the arithmetic mean across all metrics as the primary score (Table 2). Overall, GatorOnco achieved the best average performance score (Overall = 0.6002), representing a 29.7% relative improvement over Llama-3.1-8B-Instruct without domain adaptation (0.4626). GatorOnco also outperformed substantially larger models, including MedGemma-27B and Llama-3.3-70B-Instruct.  GatorOnco achieved the highest scores across most evaluation metrics. The largest performance gains were observed on semantic and faithfulness scores compared with lexical overlap-based metrics. For example, GatorOnco achieved the highest AlignScore (0.4847), exceeding Llama-3.1-8B-Instruct (0.3820) and outperforming the larger MedGemma-27B and Llama-3.3-70B models (0.4689 and 0.4723, respectively).

| Model | Lexical Measures | | | Semantic Measures | | Faithfulness Measures | | Overall |
|---|---|---|---|---|---|---|---|---|
| | BLEU | Meteor | Rouge-L | BERTScore | BLEURT | AlignScore | UniEval | |
| Llama 3.1-8B-Instruct | 0.2830 | 0.2190 | 0.1880 | 0.8388 | 0.6730 | 0.3820 | 0.6542 | 0.4626 |
| **GatorOnco** | **0.3638** | 0.3251 | **0.3552** | **0.8948** | **0.8670** | **0.4847** | **0.9105** | **0.6002** |
| MedGemma-27B-Text-it | 0.3360 | **0.3570** | 0.3170 | 0.8659 | 0.8190 | 0.4689 | 0.8715 | 0.5765 |
| Llama 3.3-70B-Instruct | 0.3580 | 0.3130 | 0.3460 | 0.8724 | 0.8220 | 0.4723 | 0.8796 | 0.5805 |

**Table 2. Comparison of GatorOnco with general-purpose Llama and medical LLM MedGemma.** Metrics are categorized into lexical (BLEU, Meteor, ROUGE-L), semantic (BERTScore, BLEURT), and faithfulness (AlignScore, UniEval) dimensions.

### GatorOnco achieved expert-level treatment planning for CRC, per the evaluation of five UF Health oncologists

We conducted a blinded statistical test with five UF Health oncologists using a test set of 79 CRC cases (Fig. 4a) to assess how well GatorOnco performed compared with professional UF Health oncologists. Five UF Health oncologists manually reviewed treatment plans generated by GatorOnco, Llama3.1-8B-Instruct, and the ground-truth treatment plans reviewed by oncologists. The treatment plans were randomly mixed with ground truth for a blind evaluation. Oncologists rated the plans using a five-point Likert scale[23] with five dimensions: correctness, currency (guideline adherence), safety, readability, and completeness (Fig. 4b). Inter-rater agreement analysis showed overall consistency across these evaluation dimensions, with the highest agreement observed for completeness ($\alpha = 0.849$) and readability ($\alpha = 0.757$), and moderate agreement for correctness ($\alpha = 0.626$), currency ($\alpha = 0.679$), and safety ($\alpha = 0.645$). A linear mixed-effects model with random intercepts for rater and case identity was used to isolate true system-level effects from rater-to-rater variation.

Per oncologists' evaluation, GatorOnco achieved the best overall scores, significantly better than other LLMs and on par with oncologist-authored plans in composite ratings (Fig. 4c; Fig. 4d). In the mixed-effects model (Score ~ System + (1|Rater) + (1|Sample)), GatorOnco showed the strongest fixed-effect estimate ($\beta = 1.064$, SE = 0.038, $P < 0.01$), followed by the expert gold standard ($\beta = 0.928$, SE = 0.038, $P < 0.01$), both significantly better than open-source Llama model ($P < 0.01$ for all pairwise contrasts). Compared with oncologist-authored ground-truth plans, GatorOnco received higher composite scores ($\beta = 0.136$, SE = 0.038, $P < 0.01$), indicating that GatorOnco achieved expert-level treatment planning for CRC.

We conducted dimension-specific analyses using the Aligned Rank Transform (ART) for deeper insights of the 5 evaluation dimensions (Fig. 4e). For correctness and currency, GatorOnco achieved batter ratings but with statistical parity with oncologist experts ($P = 0.921$ and $P = 0.478$, respectively), and the treatment plans written by GatorOnco and UF Health oncologists' experts both received significantly better ratings than the plans written by Llama ($P < 0.01$). GatorOnco also performed on par with expert oncologists for safety (4.22 versus 4.22, $P = 0.9995$), whereas the general-purpose LLM, Llama, received significantly lower ratings (4.22 versus 2.72, $P < 0.01$ for both comparisons). For readability, GatorOnco was rated significantly higher than both expert oncologists (4.46 versus 4.19, $P < 0.01$) and the Llama baseline (4.46 versus 4.22, $P < 0.01$). Oncologists agree that AI-generated treatment plans were consistently organized and easier to read and digest than the ground-truth treatment plans documented in the clinical notes. This is not surprising, as many studies have reported similar findings that generative AI has better readability[24,25]. For the completeness measure, our GatorOnco achieved better scores than oncologists (3.91 versus 3.52, P < 0.01), whereas the difference between GatorOnco and the Llama baseline was not statistically significant (3.91 versus 3.77, P = 0.079).

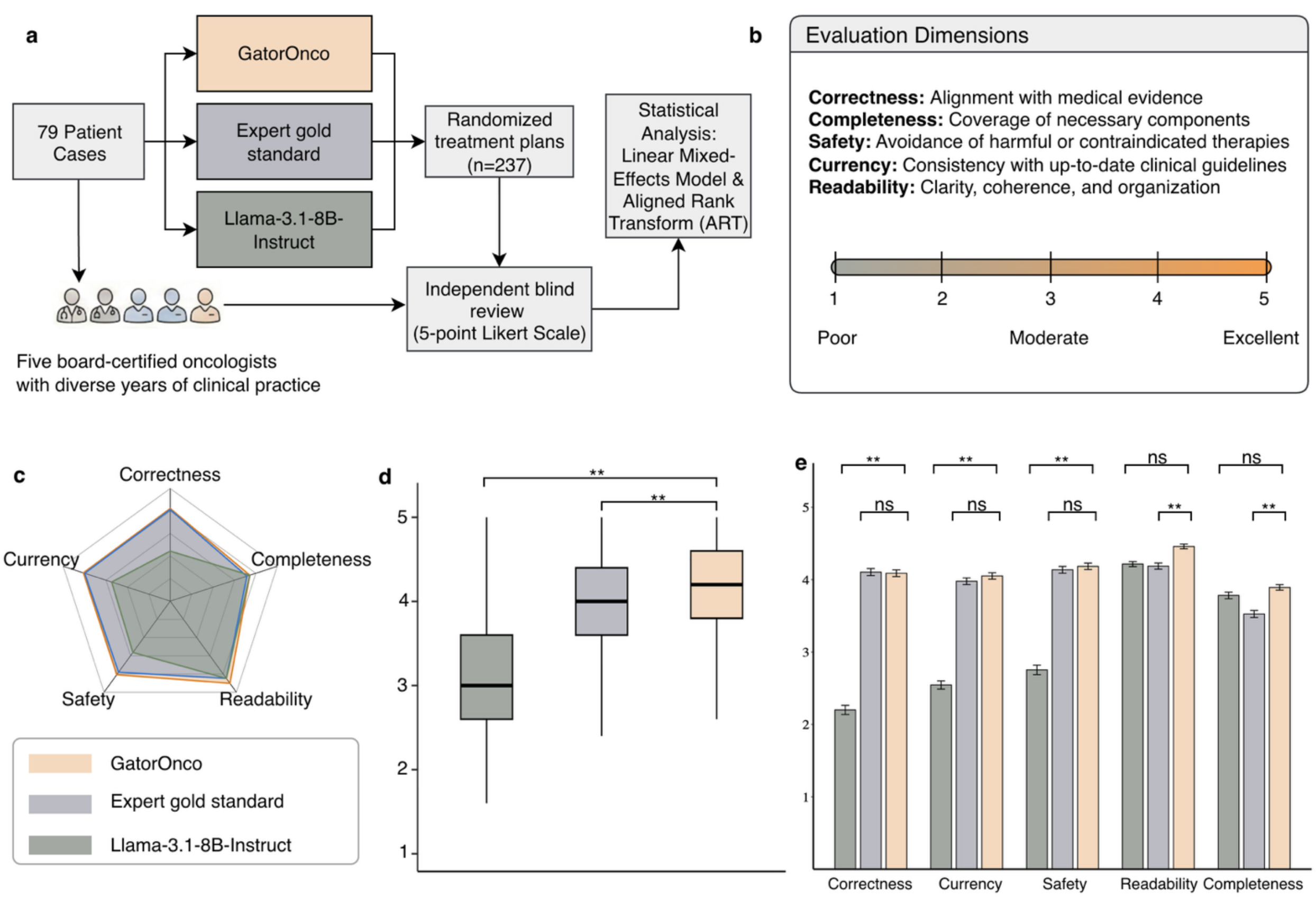

**Fig. 4: Human evaluation of model-generated treatment plans.**

**a,** Schematic of the blinded, randomized evaluation across 79 patient cases. **b,** Clinical evaluation dimensions and 5-point scoring scale. **c,** Dimension-specific scores for correctness, currency, safety, readability, and completeness. **d,** Composite score distribution. **P < 0.01 (linear mixed-effects models). E. Mean scores by individual dimension. Error bars, s.e.m.; **, P < 0.01; *, P < 0.05; ns, not significant. (ART analysis).

## Discussion

We present GatorOnco, an agentic generative AI for CRC treatment planning. We developed GatorOnco using the UF Health system-scale EHRs and applied pretraining, model merging, post-training, reinforcement learning, and agentic-based RAG to instruct GatorOnco to learn from UF Health oncologists' reasoning logic and treatment planning.

We demonstrate that general-purpose LLMs need to be adapted to the clinical context for high-stakes clinical tasks such as treatment planning, and that the deep involvement of oncologists at every step, from training to evaluation, is critical. While LLMs have transformed medical question answering and information extraction, their adoption for personalized treatment planning—the cognitive apex of oncology—has been jeopardized by technical barriers such as catastrophic forgetting and safety barriers, such as hallucinations[26–28]. GatorOnco alleviates these fundamental barriers through a novel synergy of model merging, agentic reinforcement learning, and hierarchical clinical rewards, and comprehensive involvement of oncologists. By deeply integrating oncologists into real-world case construction, loss function design, training, and evaluation, we show that an 8-billion-parameter domain-adapted local LLM can significantly outperform much larger general-purpose models (e.g., 70 billion). This finding suggests that for specialized, high-stakes healthcare tasks, data quality and domain alignment are more critical than the size of AI models. Without proper adaptation using domain-specific knowledge and data, the widely used general-purpose LLMs are not on par with healthcare professionals. Scale-up is not a silver bullet.

A well-known challenge in adopting general-purpose open-source LLMs to the clinical domain is the "forgetting issue", where LLMs lose instruction-following capabilities after fine-tuning on specialized corpora[29,30]. We effectively circumvented this by employing a model-merging technique[20] to synthesize clinical knowledge with instruction-following capabilities. GatorOnco achieved expert-level performance indistinguishable from that of UF Health oncologists in terms of correctness and safety, whereas the unadopted open-source LLMs failed. Our findings show that it's necessary for general-purpose LLMs to undergo rigorous domain adaptation using real-world clinical data (e.g., our 166B clinical tokens) and clinicians' knowledge to ensure safety[31]. This study contributes a generalizable solution for adapting open-source models for high-stakes clinical applications.

Hallucinations remain a challenge for generative AI in medicine. We demonstrate that agentic reasoning, grounded in retrieval-augmented generation (RAG), offers a robust solution[27]. Leveraging the reasoning power, GatorOnco's explicit reasoning traces allow clinicians to audit the decision-making process, linking recommendations directly to retrieved NCCN guidelines and patient-specific evidence. This transparency is crucial for clinicians' trust in AI tools. All AI models received high ratings for readability ($P < 0.01$), reflecting the success of LLMs in linguistic ability. However, real-world data and deep clinician involvement are needed to ensure clinical validity and safety.

We discover a bottleneck in developing AI for medicine: the lack of a machine-friendly format of clinical guidelines. While we successfully engineered an agentic RAG solution, the current structure of guidelines (e.g., static PDFs) is inherently unfriendly to machines[32]. Significant manual effort was required to parse eligibility criteria and treatment pathways. As AI has increasingly been used in healthcare, we advocate a paradigm shift toward "AI-friendly" clinical guideline releases, such as the Fast Healthcare Interoperability Resources (FHIR) standard. Professional bodies should publish guidelines with machine-readable schemas, encoded therapeutic intents, and explicit change logs. Such infrastructure would accelerate the deployment of concurrency-aware healthcare AI systems.

The evaluation of generative AI for clinical applications remains challenging[33]. Automatically generated machine-based metrics heavily rely on surface-level text similarity and largely do not reflect real-world clinicians. However, human evaluation using healthcare professionals requires continuous commitment from healthcare providers[34].

This study has limitations. Our findings are derived from a single disease domain and from a single institution. The note structures, documentation conventions, and data quality may differ across health systems. For instance, the specific templates and workflows intrinsic to UF Health's Epic implementation may not seamlessly map to other institutional infrastructures. Second, we observed that performance fluctuates with prompting structure; while we harmonized inputs for this study (see Supplementary Table 3), developing robust prompt-agnostic systems remains a challenge for generative AI. GatorOnco achieved expert-level treatment planning in a blind test using 79 cases, yet that doesn't indicate it's mature enough for deployment in cancer care. GatorOnco remains a prototype tool that could assist oncologists with treatment planning for CRC. Many challenging steps are needed to make GatorOnco a practical tool for real-world cancer care, including, but not limited to, a fail-safe mechanism, uncertainty quantification, deferral policies when GatorOnco has low confidence, and integration with existing clinical workflows. We will further improve GatorOnco on more populations and cancer types and seek to test its efficacy through AI-related trials.

## Methods

### Datasets

### Data acquisition and corpus construction

**Clinical narratives corpus.** This study used clinical narratives from UF Health Integrated Data Repository (IDR), the research data warehouse of UF Health. The study was approved by the UF Institutional Review Board (IRB202100049, IRB201902362). UF Health IDR team collected clinical notes from 2011 to 2025, from over 60 million encounters, 126 departments, from inpatient, outpatient, and emergency settings, yielding a proprietary corpus of 166 billion tokens. Notes were processed using a de-identification tool to remove PHI (Protected Health Information) defined by HIPAA. Patient identifiers were replaced using IRB-specific identifiers. To prevent data leakage, we excluded all clinical notes associated with the colorectal cancer cohort using the IRB-specific patient identifiers.

**Biomedical corpus construction for pre-training.** We merged the UF Health clinical corpus with additional public corpora, including FineWeb (56 billion general-domain English tokens)[35], MIMIC-III,[36] The Pile[37] (NIH, PubMed Central, PubMed Abstracts), Meditron Guideline[38], Healix-Shot[39], to generate a corpus with 282 billion tokens.

### Cohort curation and expert annotation

**UF Health CRC cohort.** We identified a cohort of CRC from the UF Health IDR by linking EHRs to the tumor registry, covering encounters between January 2012 and March 2025. Inclusion criteria were: (1) adults aged ≥18 years; (2) confirmed primary CRC diagnosis via ICD diagnosis codes and tumor registry records; and (3) had at least one clinical note within a ±1-year diagnostic window. We collected all clinical notes and filtered out duplicates and those with <1,000 tokens. Then, a group of three board-certified UF Health oncologists identified keywords to retrieve initial clinical encounter notes containing the treatment plan sections. Oncologists manually reviewed and verified the notes for this study.

**Expert annotation and quality control.** The three oncologists manually reviewed the identified notes to (1) determine if it's the initial encounter note with oncologists that contains diagnostic synthesis and initial treatment planning, (2) if yes, annotate the potential treatment options that can be considered for the patient, the treatment plan sections, and the evidence used by oncologists to write the treatment plan. This process generated a total of 395 unique CRC cases.

We implemented a multi-stage annotation procedure. The three oncologists first drafted the initial annotation guidelines, which were refined in calibration phases. Each case was independently annotated by at least two oncologists. For each case, experts annotated clinical evidence (e.g., TNM staging, biomarkers, comorbidities) and treatment plan sections. When the treatment plan is incomplete or contains clear errors, we ask oncologists to complete or modify it as needed. Disagreements were adjudicated through consensus sessions. If the consensus cannot be achieved among the three oncologists, a senior oncologist, TG, will make the final decision as an independent and blinded judge. Inter-annotator reliability was assessed using Cohen's Kappa[40] statistic. The final 395-case cohort was partitioned into training (n = 276), validation (n = 40), and held-out test (n = 79) sets using stratified sampling.

**Construct semantic vectors to index NCCN guidelines to help reasoning**

To improve LLMs' adherence to clinical guidelines, we integrated the NCCN Guidelines in Oncology for colon cancer (Version 4.2025) and rectal cancer (Version 3.2025) into the reasoning process using a hybrid retrieval strategy that combines structured metadata filtering with semantic search. Guideline text was parsed from NCCN guidelines in PDF files, and multi-branch decision workflows were converted into stepwise rules to preserve both narrative recommendations and branch-based treatment logic. The resulting corpus was segmented into semantically coherent chunks of 256 ± 50 tokens, aligned to paragraph boundaries, and embedded using the E5-large-v2[41] model in a FAISS[42] vector database. Each chunk was further annotated with oncologist-defined key data elements for treatment planning, including cancer type, stage group, metastatic status, metastatic site, and MSI/MMR status; initial annotation was assisted by ChatGPT-5 and subsequently reviewed by two oncologists.

**Post-training data construction**

We assembled a comprehensive post-training corpus comprising three complementary resources: a general clinical reasoning dataset (MedVLThinker-m23k; ~23.5k examples), a CRC-specific reasoning dataset derived from 276 UF Health training cases, and a guideline factual knowledge subset composed of direct question–answer pairs distilled from NCCN colorectal cancer guidelines.

For the CRC-specific dataset, each training case was prepared in two alternative formats for reasoning distillation: (1) raw de-identified clinical notes and (2) a narrative patient profile constructed from oncologist-reviewed clinical evidence. Each format was paired with an expert-verified treatment plan, and the resulting note/report-plus-plan pairs were used as inputs to GPT-oss-120B[43] for stepwise reasoning-trajectory distillation. All distilled trajectories and associated case annotations were reviewed and corrected as needed by oncologists. Because the source records spanned 2012–2025, oncologists also reviewed and updated historical treatment plans as needed to ensure concordance with contemporaneous NCCN standards before distillation of reasoning.

## Model architecture and training pipeline

### Pre-training and post-training to develop GatorOnco using a hierarchical reward

We adopted the pre-trained weights from Meta-Llama-3.1-8B[18], a widely used open-source LLM. We performed continual pre-training using our 282-billion-token biomedical corpus to create GatorTronLlama. To address forgetting and retain instruction-following ability, we merged this foundation model with its instruction-tuned counterpart (Meta-Llama-3.1-8B-Instruct) using an evolutionary model-merging framework (mergekit-evo). Using this merged model, we applied a two-stage post-training approach to guide treatment planning. First, we used Supervised Fine-Tuning (SFT) to fine-tune the model on more than 20,000 medical reasoning QA pairs using the default GPT-3 loss to establish core clinical writing competency. Second, we implemented an agentic Retrieval-Augmented Generation (RAG) framework to further optimize our model with Group Relative Policy Optimization (GRPO). Specifically, the agent dynamically retrieves context from the indexed NCCN guidelines to guide the reinforcement learning (RL). We worked with UF Health oncologists to design a hierarchical clinical reward function that mirrors oncological decision-making priorities. A normalized reward is first calculated based on the correctness of high-level treatment options (e.g., Chemotherapy vs. Immunotherapy). Then, a conditional reward for granular regimen (e.g., FOLFOX vs. CAPOX) is computed – if the high-level treatment option is correctly inferred; or set as zero – if the high-level treatment options are not correctly inferred. We designed this hierarchical reward to control "hallucinations" (i.e., still generating wrong treatment plans when the treatment option is not correctly inferred) to enforce a safety-first planning strategy.

**Agentic reasoning and dynamic retrieval.** To address the limitation of handling time-sensitive content in previous non-agentic generative LLMs, we adopted a dynamic, agentic reasoning model implemented in the Search-R1[44] framework. This enables GatorOnco to approach clinical treatment planning as a multi-turn Markov Decision Process (MDP), rather than a single-time text generation task. Specifically, the algorithm operates through an iterative cycle of internal reasoning and external information acquisition. Upon receiving a patient, GatorOnco initiates a "Chain-of-Thought" (CoT) process (enclosed in <think> tags) that decomposes the complex clinical scenario into sub-problems. When the agent identifies an information gap, it autonomously generates a structured search token(<search>). This action triggers the environment to pause generation, query the indexed NCCN guidelines, and append the top-ranked relevant guideline chunks to the context window. Then, the agent resumes generation by leveraging the retrieved evidence to verify, modify, or correct its initial reasoning logics before generating the treatment plan. This agentic architecture fundamentally differs from standard RAG by allowing the model to actively and dynamically decide when to search and how to use the information, closely simulating the workflow of an oncologist who consults the guidelines during decision-making.

## Optimization Objective and Hierarchical Reward

**Group relative policy optimization (GRPO)**[21] is an RL algorithm that uses the average reward of multiple sampled outputs. For a query $q$, the model generates a group of outputs $\{o_i\}_{i=1}^{G}$ from the old policy $\pi_{\theta_{old}}$. The optimization objective is defined as:

$$\mathcal{J}_{\text{GRPO}}(\theta) = \mathbb{E}_{q \sim P(Q), \{o_i\}_{i=1}^{G} \sim \pi_{\theta_{\text{old}}}(O|q)} \left[ \frac{1}{G} \sum_{i=1}^{G} \frac{1}{|o_i|} \sum_{t=1}^{|o_i|} \left( \min\left( \frac{\pi_\theta(o_{i,t} | q, o_{i,<t})}{\pi_{\theta_{\text{old}}}(o_{i,t} | q, o_{i,<t})} \widehat{A}_i, \text{clip}\left( \frac{\pi_\theta(o_{i,t} | q, o_{i,<t})}{\pi_{\theta_{\text{old}}}(o_{i,t} | q, o_{i,<t})}, 1-\varepsilon, 1+\varepsilon \right) \widehat{A}_i \right) - \beta D_{\text{KL}}(\pi_\theta \| \pi_{\text{ref}}) \right) \right]$$

where $\hat{A}_i$ is the advantage score, calculated by standardizing the rewards within the group:

$$\widehat{A}_i = \frac{R_i - \text{mean}(\{R_j\})}{\text{std}(\{R_j\})} .$$

**Hierarchical oncology-oriented reward.** We designed a reward function that first prioritizes the correct therapeutic drug regimen, then accounts for the detailed plans within the regimen to ensure clinical validity. We model the treatment evaluation as a hierarchical process conditioned on a **hard constraint ($\boldsymbol{G}$)**: the agent receives rewards for regimen selection *only if* the predicted modality (e.g., Neoadjuvant Chemotherapy) matches the expert annotation.

For the regimen-level evaluation, standard n-gram matching is insufficient due to the complex nomenclature of oncology (e.g., "CapeOx" implies "Capecitabine" and "Oxaliplatin"). Therefore, we implemented an **entity-level set-theoretic alignment**. Let $\mathcal{E}(r)$ denote the set of normalized pharmacologic entities extracted from a regimen description $r$. The semantic alignment score is calculated using the Dice similarity coefficient over these entity sets. The total reward $R$ is formalized as:

$$R = G(m_{pred} = m_{gold}) \cdot \left[ \omega_1 + \omega_2 \cdot \frac{2 \cdot |\mathcal{E}(r_{pred}) \cap \mathcal{E}(r_{gold})|}{|\mathcal{E}(r_{pred})| + |\mathcal{E}(r_{gold})|} \right]$$

This formulation ($\omega_1 = 1.0, \omega_2 = 0.5$) incentivizes the model to accurately retrieve and reason about the constituent drugs of a regimen, rewarding partial correctness (e.g., identifying 2 out of 3 drugs) rather than a strict exact match for all, inspired by the logic of clinical board examinations.

## Evaluation

**Experimental setup and baselines.** We evaluate GatorOnco and compare it with three widely used generative LLMs, including (1) Llama-3.1-8B-Instruct, a parameter-matched general-purpose model used to assess the contribution of domain adaptation; (2) MedGemma-27B-Text-it, a clinically instruction-tuned model serving as a domain-specific model; and (3) Llama-3.3-70B-Instruct, a substantially larger model to evaluate whether a smaller agentic model can outperform larger non-agentic LLMs. All LLMs were evaluated using the same prompts and experimental settings to ensure fair comparisons.

### Automatic evaluation tasks and metrics

We evaluate GatorOnco on two deterministic AI tasks, including (1) determining the high-level treatment categories, and (2) generating narrative treatment plans with oncological details.

**Determining the high-level treatments.** We evaluated the models' ability to identify appropriate therapeutic interventions across five NCCN-derived categories: surgery, radiotherapy, chemotherapy, targeted therapy, and palliative care. Performance was measured using two protocols: (1) Per-option classification F1-Score, assessing the discriminative power for individual modalities; and (2) Treatment set prediction, utilizing an "exact-match" metric where success was defined as correctly reproducing the entire set of oncologist-annotated multiple potential interventions that could be considered for a given patient.

**Generating narrative treatment plan sections.** The textual quality of generated plans was evaluated using a multidimensional framework. Lexical fidelity was measured via BLEU[45], ROUGE-L[46], and METEOR[47] to assess structural overlap with expert references. Semantic correspondence was evaluated using BERTScore[48] and BLEURT[49] to capture high-level meaning preservation beyond surface phrasing. Given the low error tolerance nature of clinical oncology, factual faithfulness was assessed using AlignScore[50] and UniEval[51], which quantify the extent to which the generated text is supported by the source input and free from hallucinations.

**Blind evaluation by UF Health Oncologists**

**Study design and participants.** We applied generative LLMs to generate treatment plans for the 79 cases in the test set. We randomly mixed the treatment plans generated by two generative LLMs, including GatorOnco and (2) Meta-Llama-3.1-8B-Instruct, with the ground truth by oncologists, resulting in a total of 237 plans. A total of 5 UF Health oncologists independently reviewed treatment plans. The raters were unaware of whether the reviewed plans were generated by AI models or written by oncologists for a blind statistical test.

**Evaluation instrument and statistical analysis.** Raters (i.e., oncologists) scored each treatment plan using a 5-point Likert scale[23] (1 = worst, 5 = best) across five dimensions: Correctness, Currency, Safety, Readability, and Completeness (see Supplementary Table 1). To compare system performance while accounting for repeated measures across raters and cases, scores were analyzed using a linear mixed-effects model with rater and case included as random intercepts (Score ~ System + (1|Rater) + (1|Sample)). Because Likert ratings are ordinal, dimension-

specific effects were further evaluated using the Aligned Rank Transform[52] (ART) procedure, which enables non-parametric inference for factorial designs. When overall system effects were significant, post-hoc pairwise comparisons were performed using ART-adjusted Wilcoxon signed-rank tests with Holm–Bonferroni correction. All tests were two-sided with a significance threshold of $\alpha < 0.05$. To assess inter-rater agreement among the five oncologists, we calculated Krippendorff's alpha for each evaluation dimension using the 237 rated plans.

**Ethics approval and consent to participate**

This study was approved by the University of Florida Institutional Review Board (IRB202100049 and IRB201902362). The study used retrospective, de-identified clinical data from UF Health. The requirement for informed consent was waived by the Institutional Review Board because the study involved retrospective analysis of de-identified data and posed minimal risk to participants.

**Data availability**

The benchmark datasets that support the findings of this study are available from the official websites of natural language processing challenges with Data Use Agreements. More specifically:

Pre-training dataset:

1. FineWeb: https://huggingface.co/datasets/HuggingFaceFW/fineweb
2. The Pile dataset: https://pile.eleuther.ai/
3. Healix-Shot: https://huggingface.co/datasets/health360/Healix-Shot
4. Meditron clinical guidelines: https://github.com/epfLLM/meditron
5. MIMIC-III clinical notes: https://physionet.org/content/mimiciii/1.4/
6. UF Health IDR clinical notes are not open to the public due to patient privacy information. We will release 79 deidentified test sets used in this study through GitHub.

SFT datasets:

1. General medical reasoning dataset: https://huggingface.co/datasets/UCSC-VLAA/MedVLThinker-m23k-tokenized

2. Guideline-derived QA pairs will be further released on Huggingface Datasets

**Computer code**

The computer codes for GatorOnco are available from:

https://github.com/uf-hobi-informatics-lab/GatorOnco

The computer codes to train GatorTronGPT models are available from:
https://github.com/NVIDIA-NeMo/NeMo/blob/main/tests/collections/llm/llama3_pretraining.py

The scripts used for model merging are available from:

https://github.com/arcee-ai/mergekit

The scripts used for agentic reinforcement learning and inference are available from:

https://github.com/PeterGriffinJin/Search-R1

## Acknowledgments

**Funding**: This study was partially supported by grants from the Patient-Centered Outcomes Research Institute® (PCORI®) Award ME-2018C3-14754 and ME-2023C3-35934, the PARADIGM program awarded by the Advanced Research Projects Agency for Health (ARPA-H), National Institute on Aging U24AG098157, National Institute of Allergy and Infectious Diseases, NIAID R01AI172875, National Heart, Lung, and Blood Institute, R01HL169277, R01HL176844, and the UF Clinical and Translational Science Institute. The content is solely the responsibility of the authors and does not necessarily represent the official views of the funding institutions.

**Support from UF Research Computing:** We would like to thank the UF Research Computing team, led by Dr. Erik Deumens, for providing computing power through UF HiperGator-AI cluster.

We thank Xiaohan Li, Jinqian Pan and Kai Zhang for their assistance with code refinement and software engineering support for this study.

## Author contributions

YW, TG, JB, YG, CP, and ML were responsible for the overall design, development, and evaluation of this study. ML, CP, ZC, and MZ had full access to all the data in the study and took responsibility for the integrity of the data and the accuracy of the data analysis. YG and YW designed the blind statistical test of the treatment planning performance of GatorOnco. AL performed all statistical analyses. TJ, LE, TL, and TG are oncologists who were involved in patient case review and annotation. TJ, TL, LG, CS, and OM are the oncologists who performed the human evaluation. YW, ML, CP, and YG did the bulk of the writing, and EAS, DAM, TM, GL, and LS also contributed to the writing and editing of this manuscript. KS contributed to study implementation support and coordination. YZ provided research computing resources and infrastructure support. All authors reviewed the manuscript critically for scientific content, and all authors gave final approval of the manuscript for publication.

**Competing interests**

The Authors declare no Competing Financial or Non-Financial Interests.

**Materials & Correspondence**: Yonghui Wu, PhD

**Statistical information: N/A**